\documentclass[10pt,twocolumn,letterpaper]{article}

\usepackage{cvpr}              
\usepackage{comment}
\usepackage{multirow}
\usepackage{bbding}

\usepackage{bm}

\usepackage{microtype}

\definecolor{cvprblue}{rgb}{0.21,0.49,0.74}
\usepackage[pagebackref,breaklinks,colorlinks,allcolors=cvprblue]{hyperref}

\def\paperID{7333} 
\def\confName{CVPR}
\def\confYear{2025}

\title{Sketchy Bounding-box Supervision for 3D Instance Segmentation}

\author{Qian Deng\textsuperscript{\rm 1}, ~Le Hui\textsuperscript{\rm 2}, Jin Xie\textsuperscript{\rm 3,4}\thanks{Corresponding author}, ~Jian Yang\textsuperscript{\rm 1}\\
	\textsuperscript{\rm 1}PCA Lab, College of Computer Science, Nankai University, Tianjin, China\\ 
	\textsuperscript{\rm 2}School of Electronics and Information, Northwestern Polytechnical University, Xi’an, China \\
	\textsuperscript{\rm 3}State Key Laboratory for Novel Software Technology, Nanjing University, Nanjing, China \\
	\textsuperscript{\rm 4}School of Intelligence Science and Technology, Nanjing University, Suzhou, China \\
	\tt\small dengqian@mail.nankai.edu.cn; huile@nwpu.edu.cn;  csjxie@nju.edu.cn; csjyang@nankai.edu.cn
}

\begin{document}
\maketitle
\begin{abstract}
Bounding box supervision has gained considerable attention in weakly supervised 3D instance segmentation. While this approach alleviates the need for extensive point-level annotations, obtaining accurate bounding boxes in practical applications remains challenging. To this end, we explore the inaccurate bounding box, named sketchy bounding box, which is imitated through perturbing ground truth bounding box by adding scaling, translation, and rotation. In this paper, we propose Sketchy-3DIS, a novel weakly 3D instance segmentation framework, which jointly learns pseudo labeler and segmentator to improve the performance under the sketchy bounding-box supervisions. Specifically, we first propose an adaptive box-to-point pseudo labeler that adaptively learns to assign points located in the overlapped parts between two sketchy bounding boxes to the correct instance, resulting in compact and pure pseudo instance labels. Then, we present a coarse-to-fine instance segmentator that first predicts coarse instances from the entire point cloud and then learns fine instances based on the region of coarse instances. Finally, by using the pseudo instance labels to supervise the instance segmentator, we can gradually generate high-quality instances through joint training. Extensive experiments show that our method achieves state-of-the-art performance on both the ScanNetV2 and S3DIS benchmarks, and even outperforms several fully supervised methods using sketchy bounding boxes. Code is available at \url{https://github.com/dengq7/Sketchy-3DIS}.

\end{abstract}





\section{Introduction}
\label{sec:intro}

With the rise of 3D datasets and the growing need for 3D scene understanding, various 3D computer vision tasks~\cite{qi2019deep, xie2020mlcvnet,misra2021end, zhang2020h3dnet, liu2021group, wang2022rbgnet, zheng2022hyperdet3d, rukhovich2022fcaf3d, wang2022cagroup3d, schult2023mask3d, lai2023mask, sun2023superpoint, ngo2023isbnet, ngo2023gapro} have gained significant attention from both academia and industry.
Among these, 3D instance segmentation is both fundamental and challenging, as it involves recognizing the category of each object within a point cloud and delineating its individual mask. Current 3D instance segmentation methods~\cite{schult2023mask3d, sun2023superpoint, ngo2023isbnet, hui2022learning, chen2021hierarchical, jiang2020pointgroup, vu2022softgroup} have demonstrated impressive performance, primarily rely on data-hungry deep learning approaches that require substantial training data and dense point-level manual annotations. Notably, point-level fully supervised instance annotation in ScanNet takes approximately 22.3 minutes~\cite{dai2017scannet} per scene, while labeling a 3D box in ScanNet takes about 1.93 minutes~\cite{tao2022seggroup} per scene. Therefore, bounding box annotation stands out as it provides substantial instance-level information while maintaining a reasonable annotation cost, making it a favorable annotation choice for weakly supervised 3D instance segmentation.

\begin{figure}[t]
  \centering
   \includegraphics[width=0.95\linewidth]{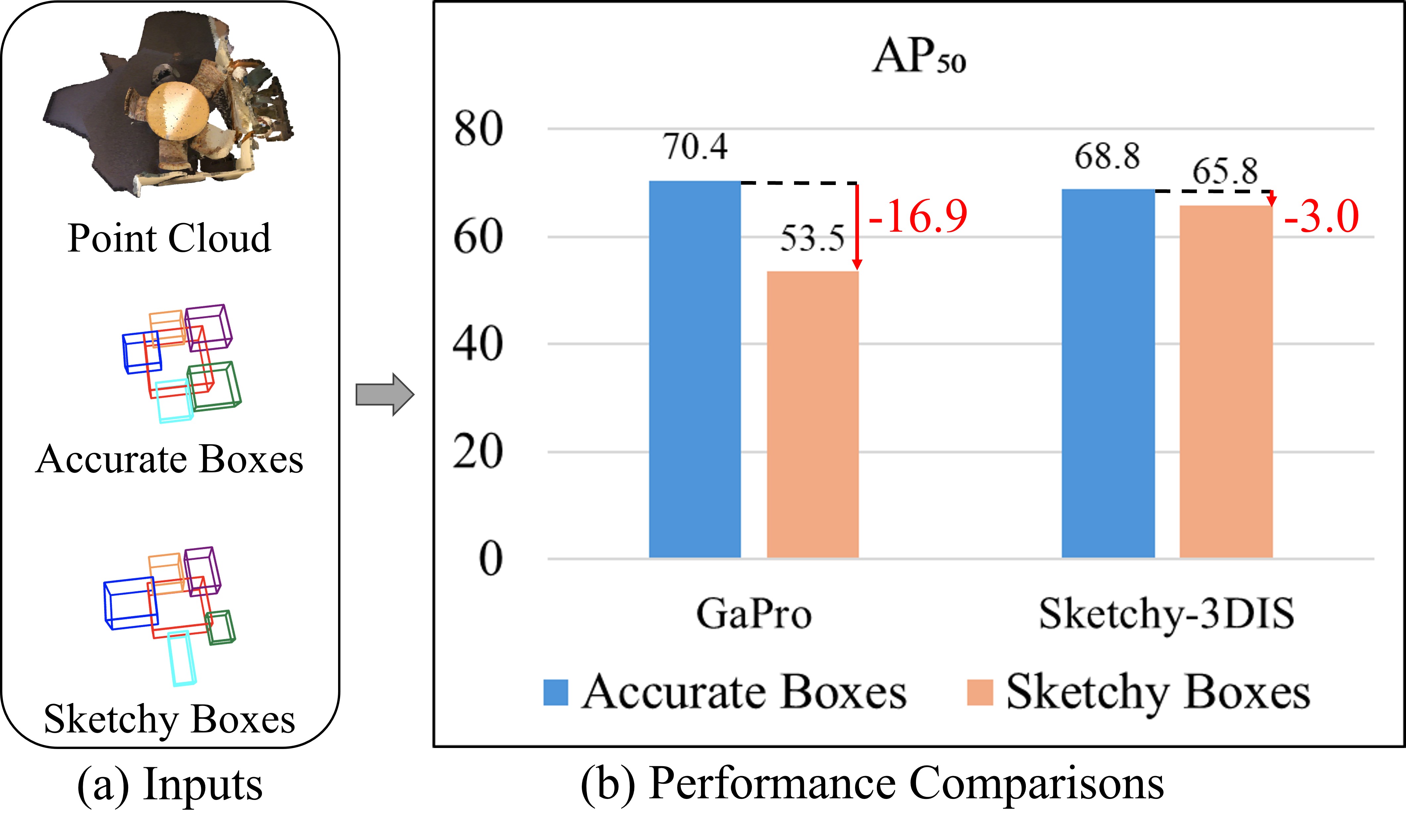}
   \vspace{-1.0em}
   \caption{(a) illustrates the inputs of bounding box supervised 3D instance segmentation. (b) compares the performances of GaPro~\cite{ngo2023gapro} and our Sketchy-3DIS under both accurate and scaled sketchy boxes supervision on ScanNetV2 validation set.}
   \label{fig:figure01}
   \vspace{-2.0em}
\end{figure}

Many weakly supervised methods have emerged to tackle the 3D instance segmentation task. Semi-supervised 3D instance segmentation methods~\cite{liao2021point, chu2022twist, xie2020pointcontrast, hou2021exploring} use training data with partial labels, while others~\cite{liao2021point, chu2022twist, xie2020pointcontrast, hou2021exploring} utilize indirect labels, such as 2D image annotations. 
Additionally, object-level annotations, like marking a single point~\cite{liu2021one, tao2022seggroup, dong2022rwseg, tang2022learning} or drawing a bounding box~\cite{chibane2022box2mask, yu20243d, ngo2023gapro, du2023weakly} for each object, are employed. Bounding box supervision is another important annotation for weakly supervised 3D instance segmentation. Box2Mask~\cite{chibane2022box2mask} is the first attempt that uses bounding box supervision for 3D instance segmentation. Subsequently, the following methods~\cite{du2023weakly,ngo2023gapro,lu2024bsnet,yu20243d} use different constraints to improve performance, such as using local correlations, Gaussian processes, teacher-student model, SAM~\cite{kirillov2023segment}, respectively. Although impressive results have been achieved in weakly supervised 3D instance segmentation, these methods still rely on accurate and compact bounding box annotations. It is worth noting that in practice, accurate and compact boxes are difficult to obtain. Generally, annotated bounding box involves changes in scaling, translation, and rotation. As shown in Fig.\ref{fig:figure01}, we observed that the performance of GaPro~\cite{ngo2023gapro} decreases significantly under scaled bounding box supervision. This motivated us to design a new approach to tackle with these inaccurate bounding box annotations.

In this paper, we propose a weakly supervised 3D instance segmentation framework to tackle sketchy bounding box supervision. To simulate sketchy bounding boxes, we impose scaling, translation, and rotation on the ground truth bounding boxes to generate inaccurate bounding boxes. Our method consists of two key components: adaptive box-to-point pseudo labeler and coarse-to-fine instance segmentator. Given a point cloud scene and sketchy bounding boxes, we first use 3D U-Net~\cite{graham20183d} backbone to extract point features. After that, we feed the learned features and sketchy bounding boxes of the scene into the adaptive box-to-point pseudo labeler to generate high-quality point-level instance labels. For the points located in the overlapped region between two sketchy bounding boxes, we adaptively assign the points to the corresponding box by learning the similarity of points and boxes. For the remaining points, we directly assign them to the corresponding box according to the spatial correlations. In this way, we can transform the sketchy bounding box into a compact bounding box. We consider the labels of points that located within the generated compact bounding box as pseudo instance labels. To accurately predict instances, we input the learned features into the coarse-to-fine instance segmentator iteratively. We first detect the coarse bounding box and mask of the instance by executing queries on the whole point cloud. Based on the small region of the coarse bounding box, we re-predict the fine bounding box and mask by conducting queries within a more precise small region. Finally, by jointly optimizing the adaptive box-to-point pseudo labeler and coarse-to-fine instance segmentator, our method can gradually improve the performance of weakly supervised 3D instance segmentation using the sketchy bounding box annotations. Rich experiments on the ScanNetV2 and S3DIS show that our method have achieved state-of-the-art performance.

Our contributions are as follows:
\begin{itemize}
\item{
We are the first to explore the use of sketchy bounding box annotations for 3D instance segmentation, which allows for training with noise-tolerant, instance-size-agnostic 3D bounding boxes.
}
\end{itemize}

\begin{itemize}
\item{
We propose a new weakly supervised 3D instance segmentation framework, which presents an adaptive box-to-point pseudo labeler and a coarse-to-fine instance segmentator for joint training.
}
\end{itemize}

\begin{itemize}
\item{
Our method achieves leading performance on both the ScanNetV2 and S3DIS datasets for weakly 3D instance segmentation under sketchy bounding box annotations. 
}
\end{itemize}

\section{Related Work}
\label{sec:relatedwork}
\subsection{Fully Supervised 3D Instance Segmentation}
\label{sec:relatedwork01}


Existing fully supervised methods for 3D instance segmentation can be broadly categorized into three main pipelines: grouping-based~\cite{jiang2020pointgroup, chen2021hierarchical, liang2021instance, hui2022learning}, detection-based~\cite{vu2022softgroup, yi2019gspn, hou20193d, yang2019learning}, and query-based~\cite{schult2023mask3d, sun2023superpoint, ngo2023isbnet, he2021dyco3d, lai2023mask, shin2024spherical, kolodiazhnyi2024oneformer3d, lu2023query, tran2024msta3d} methods. Grouping-based methods typically group points based on predicted per-point semantic classes and instance center offsets, but the grouping rules are often manually defined by researchers, which lack flexibility. Detection-based methods, on the other hand, first predict the bounding box for each object and then generate the foreground mask within the box, treating the task as a two-step process of object detection followed by segmentation. More recently, inspired by DETR~\cite{zhu2020deformable} and Mask2Former~\cite{cheng2022masked}, query-based methods have been proposed for 3D instance segmentation. These methods simultaneously predict the semantic classes and instance masks of objects using a set of learnable query vectors, which model the objects' geometrical and semantical attributes integrally. Among these methods, query-based approaches tend to achieve superior performance, as the other two pipelines both face challenges such as error accumulation and time-consuming refinement processes.
 While these fully-supervised methods have made significant progress, they all rely on dense per-point annotations, and their performance tends to degrade when such annotations are unavailable. Therefore, several weakly-supervised methods have been proposed to alleviate this problem.

\subsection{Weakly Supervised 3D Instance Segmentation}
\label{sec:relatedwork02}

Weakly-supervised methods require partial or implicit annotations. ~\cite{chu2022twist, xie2020pointcontrast, hou2021exploring, liao2021point} use sparse point annotations and design various label propagation strategies. However, these sparse points cannot fully capture the spatial extent of an object. Approaches like MIT~\cite{yang20232d} and CIP-WPIS~\cite{yu20243d} leverage 2D images as additional input for 3D instance segmentation, but this introduces multi-modal dependencies. Other methods~\cite{liu2021one, tao2022seggroup, tang2022learning, dong2022rwseg} rely on instance-level signals, where labeled a single point for per instance, and instance predictions are made using techniques like random walk or clustering. Recently, most state-of-the-art methods~\cite{chibane2022box2mask, yu20243d, ngo2023gapro, du2023weakly, lu2024bsnet} have adopted 3D bounding boxes as supervision, where a single bounding box is assigned to each object. These approaches naturally capture the spatial extent of an object and the annotated bounding boxes are easier to obtain compared to dense per-point annotations.

\textbf{Bounding-box Supervised 3D Instance Segmentation.} Bounding boxes have become a widely used supervision signal, not only for 2D instance segmentation tasks~\cite{tian2021boxinst, ke2023mask, lan2021discobox} but also for 3D instance segmentation. Box2mask~\cite{chibane2022box2mask} is a pioneering approach that utilizes bounding-box supervision for 3D instance segmentation. WISGP~\cite{du2023weakly} divides the point cloud into a determined set and an undetermined set based on explicit spatial correlations, then uses polygon meshes and superpoints to label the undetermined set. GaPro~\cite{ngo2023gapro} focuses on generating reliable pseudo labels using a Gaussian process. Recently, BSNet~\cite{lu2024bsnet} employs a simulation-assisted mean teacher framework to generate pseudo labels, enhancing data samples by synthesizing scenes that contain overlapping objects. However, these methods all rely on axis-aligned bounding boxes for supervision, and their performances heavily depend on the accuracy of these annotated boxes. To address this, CIP-WPIS~\cite{yu20243d} proposes using slightly looser bounding boxes, projecting the point cloud into 2D images, and obtaining 3D instance masks based on segmented 2D masks. While effective, it relies on 2D modality and can only adapt to slightly enlarged bounding boxes, limiting its flexibility.
To alleviate the annotation costs as well as obtain robust and effective performance, we design a novel framework Sketchy-3DIS, which endeavors to not only face the common challenge of correctly assigning the points within multiple boxes but also be dedicated to correcting the errors brought by sketchy bounding boxes.
\section{Method}
\label{sec:method}


\subsection{Overview}
\label{sec:method01}
The overall framework of our weakly supervised 3D instance segmentation based on sketchy bounding boxes is depicted in Fig.\ref{fig:figure02}. It consists of two key components: adaptive box-to-point pseudo labeler and coarse-to-fine instance segmentator.

Assuming a 3D point cloud $\bm{P}\in\mathbb{R}^{N \times C}$, where $N$ and $C$ represent the number of 3D points and channels, respectively. In the basic setting, each point is represented by the 3D coordinate, RGB color, and surface normal vector.
In the training phase, we first feed the input point cloud and the sketchy bounding boxes (defined in Sec.~\ref{sec:method02}) into the 3D UNet backbone network for feature extraction, then we average the point features located in the superpoint~\cite{landrieu2018large} to generate superpoint-level features.
Next, we input the features and sketchy bounding boxes to the adaptive box-to-point pseudo labeler (Sec.~\ref{sec:method03}) for converting the rough box-level annotation to the detailed point-level annotation of the instance. Meanwhile, we feed the feature into the coarse-to-fine instance segmentator (Sec.~\ref{sec:method04}) to predict instances by Transformer. Finally, we employ bilateral matching between the generated pseudo labels (regarded as the ground truth) and the predicted instances to build the correspondence between them for training (Sec.~\ref{sec:method05}). In the inference phase, a new point cloud is fed to the backbone network and the coarse-to-fine instance segmentator to segment instances.

\begin{figure}[t]
  \centering
   \includegraphics[width=0.95\linewidth]{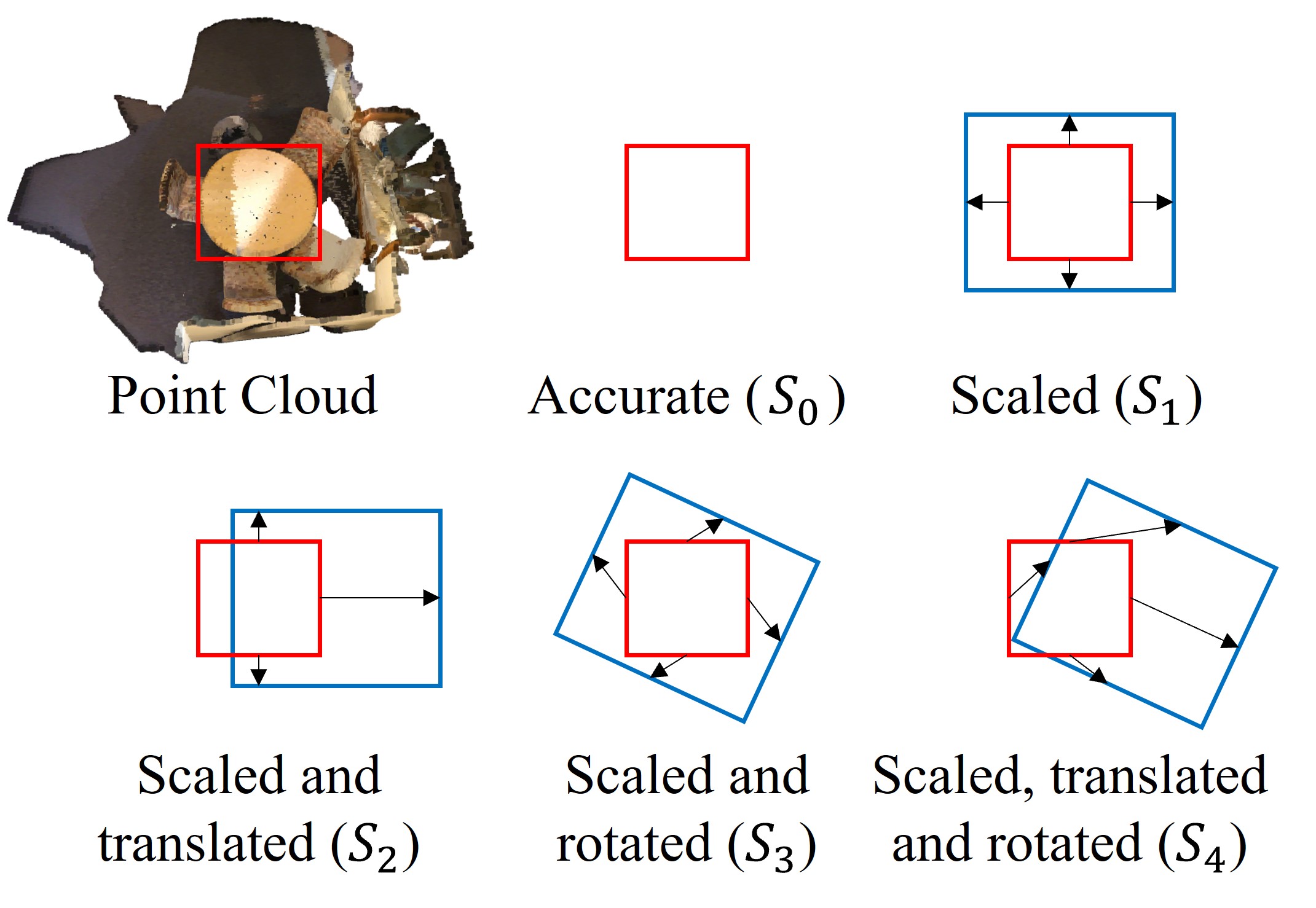}
   \vspace{-1.0em}
   \caption{Various sketchy bounding boxes under scaling, translation, and rotation perturbations.}
   \label{fig:figure10}
   \vspace{-1.0em}
\end{figure}

\begin{figure*}[t]
  \centering
   \includegraphics[width=0.95\linewidth]{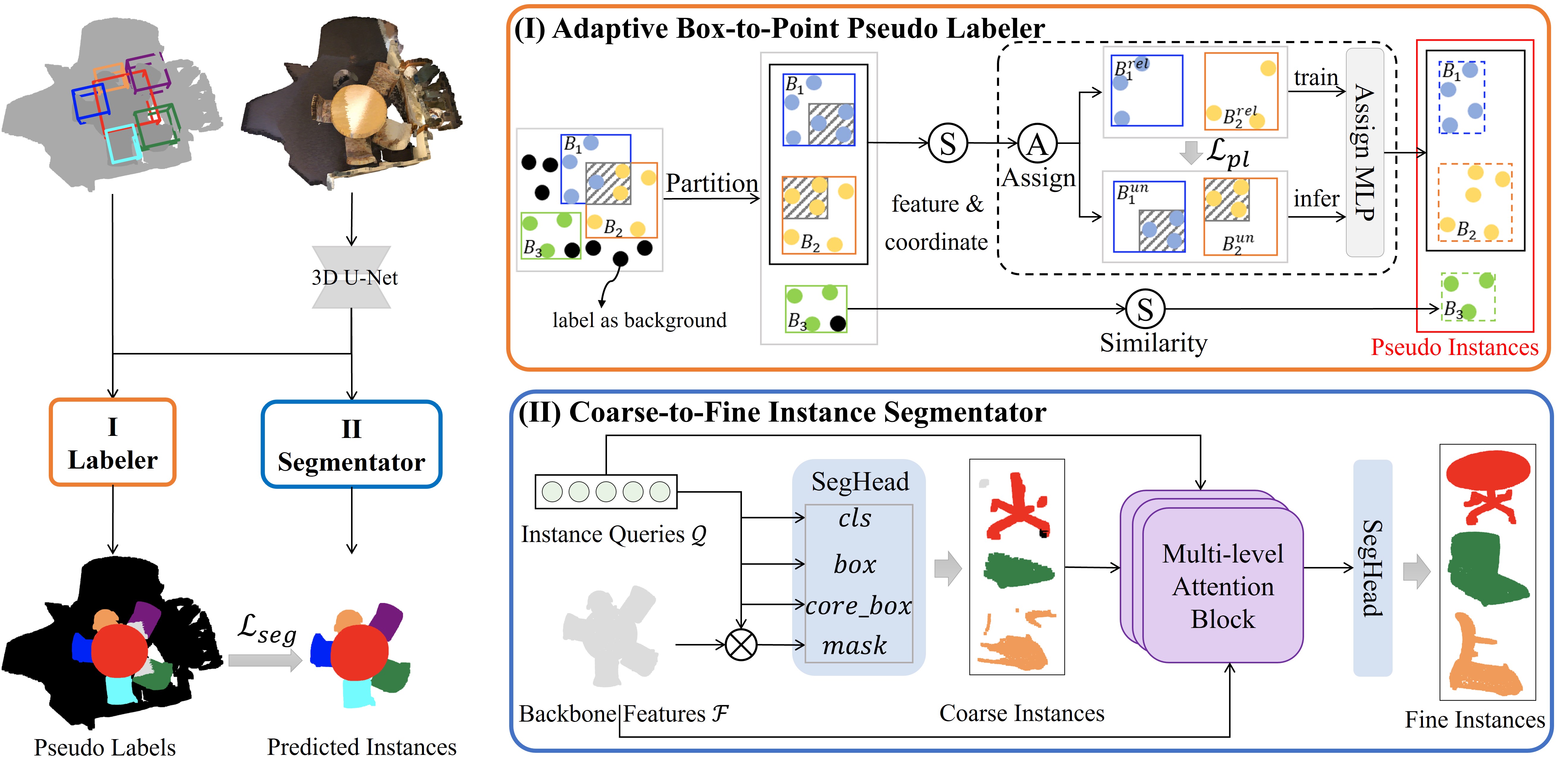}
   \vspace{-1.0em}
   \caption{\textbf{The framework of the Sketchy-3DIS.} Given a point cloud with sketchy bounding-box annotations, we first extract the backbone features using a 3D U-Net backbone, then feed them into the adaptive box-to-point pseudo labeler and the coarse-to-fine instance segmentator, finally, we utilize the generated high quality pseudo labels to supervise the predicted instances periodically.}
   \label{fig:figure02}
   \vspace{-1.0em}
\end{figure*}

\subsection{Sketchy Bounding Box Setting}
\label{sec:method02}
Our method focuses on studying weakly supervised 3D instance segmentation based on the sketchy bounding box, $i.e.$, inaccurate bounding box. In practice, it is unable to acquire compact 3D bounding box annotations for weakly supervised 3D instance segmentation. Therefore, we perturb the ground truth bounding box with scaling, translation and rotation to mimic the sketchy bounding boxes. As shown in Fig.\ref{fig:figure10}, we illustrate the generation process of different types of sketchy bounding boxes.

Specifically, given the ground truth bounding box, which is denoted by the position of two corner points:
\begin{equation}
\setlength{\abovedisplayskip}{2pt}
\setlength{\belowdisplayskip}{2pt}
    \begin{aligned}
     \mathbf{B} & = [\mathit{B}_{min}, \mathit{B}_{max}] \\
     & =[x_{min}, y_{min}, z_{min}, x_{max}, y_{max}, z_{max}]
      \label{eq:eq32_01}
    \end{aligned}
\end{equation}
First, we individually employ scaling, translation, and rotation on the ground truth bounding box to generate the basic inaccurate bounding boxes, which are formulated as:
\begin{equation}
\setlength{\abovedisplayskip}{2pt}
\setlength{\belowdisplayskip}{2pt}
\begin{aligned}
    &\mathbf{B_{scaled}} = [\mathit{B}_{min}-\alpha E, \mathit{B}_{max}+\alpha E] \\
    &\mathbf{B_{translated}} = [\mathit{B}_{min}+\beta E, \mathit{B}_{max}+\beta E] \\
    &\mathbf{B_{rotated}} = r([\mathit{B}_{min}, \mathit{B}_{max}], ME, \gamma\pi/180)
    \label{eq:eq32_02}
\end{aligned}
\end{equation}
where $\alpha$, $\beta$, $\gamma$ are the parameters of different operations. Note that $E$=$\mathit{B}_{max}$-$\mathit{B}_{min}$ and $ME$=$(\mathit{B}_{max}$+$\mathit{B}_{min})/2$. Considering that manually annotated 3D bounding boxes will not have significant differences from compact 3D ground truth bounding boxes. Therefore, we empirically set small values for those parameters. In experiments, we set $\alpha$=5\%, $\beta$=5\%, and $\gamma$=5, respectively. Then, we generate diverse sketchy bounding boxes by combining these basic operations $\mathbf{B_{scaled}}$, $\mathbf{B_{translated}}$, and $\mathbf{B_{rotated}}$. In Fig.\ref{fig:figure10}, we show four types of sketchy bounding boxes, named $S_{1}$, $S_{2}$, $S_{3}$, and $S_{4}$, the red rectangle denotes the accurate box and the blue rectangles denote various sketchy bounding boxes. It can be observed that the ``sketchy degree" increases from $S_{1}$ to $S_{4}$, making instance segmentation more challenging.



\subsection{Adaptive Box-to-Point Pseudo Labeler}
\label{sec:method03}


The sketchy 3D bounding box supervision is an inaccurate weak annotation for 3D instance segmentation. In order to convert rough annotated boxes into fine-grained point-level instance annotations, we propose an adaptive box-to-point pseudo labeler that can assign the points located in the box to the corresponding instance. Specifically, we first distinguish the points that lie in bounding boxes, and then adaptively assign them to the corresponding instance by learning the correlation between the points and 3D bounding boxes.

\textbf{Box Conditioned Points Partition.} According to the relationship between points and 3D bounding boxes, there are three types of points: the points located outside the bounding boxes ($i.e.$, background points), the points lying in a single bounding box, and the points belonging to multiple bounding boxes. Since the background points (see black circle points in Fig.~\ref{fig:figure02}) are not within any 3D bounding boxes, they are directly labeled as background. The point lying in a single or multiple bounding box could be a point on the target (see green/yellow/blue circle points in Fig.~\ref{fig:figure02}) or a background point. Therefore, we need other strategies to help distinguish the target points from the background points.

\textbf{Adaptive Point-to-Instance Assignment.} For the points lying in a single 3D bounding box (see bounding box $B_3$ in Fig.~\ref{fig:figure02}), we filter the background points ($i.e.$, black circle points in $B_3$) by comparing the similarity between the points and the 3D bounding box. Specifically, we consider the similarity in both coordinate space and feature space. The box coordinate and box feature is obtained by averaging the points' coordinates and features within them. The similarity between the points and the 3D bounding boxes can be formulated as follows:
\begin{equation}
\setlength{\abovedisplayskip}{2pt}
\setlength{\belowdisplayskip}{2pt}
    \bm{s}_{p,B} = \operatorname{cos}(\bm{f}_{p},
     \bm{f}_{B}) \times e^{-|\bm{c}_{B}
     \bm{c}_{p}|}
    \label{eq:eq01}
\end{equation}
where $\operatorname{cos}(\cdot,\cdot)$ denotes cosine similarity. In addition, $\bm{f}_{p}$ and $\bm{f}_{B}$ mean the features of point and box, while $\bm{c}_p$ and $\bm{c}_B$ mean the coordinates of point and box. In this way, it is expected to filter background points with low similarity.

For the points lying in the overlapped parts of multiple 3D bounding boxes (usually two boxes, see boxes $B_1$ and $B_2$ in Fig.~\ref{fig:figure02}), they are assigned to the corresponding instances by learning the similarity between the points and the bounding boxes. However, due to the uncertainty of point labels in overlapping areas, it is difficult to obtain high-confidence semantic labels for these points from the bounding boxes. Therefore, we consider removing the overlapping points within each box and using the remaining reliable points (see $B_1^{rel}$ and $B_2^{rel}$ in Fig.~\ref{fig:figure02}) for similarity learning. Since reliable points are only located within one box, their labels are consistent with the box labels. Given a point feature $\bm{f}_{p}$, box features  $\bm{f}_{B_{1}^{rel}}$ and $\bm{f}_{B_{2}^{rel}}$, we utilize a multilayer perceptron (MLP) network to learn the similarity score $\bm{A}_{p2b}$, which is given by:
\begin{equation}
\setlength{\abovedisplayskip}{2pt}
\setlength{\belowdisplayskip}{2pt}
 \bm{A}_{p2b} = \operatorname{MLP}(\bm{f}_{p}, \bm{f}_{p}-\bm{f}_{B_{1}^{rel}}, \bm{f}_{p}-\bm{f}_{B_{2}^{rel}})
  \label{eq:eq02} 
\end{equation}
where $\bm{f}_{p}-\bm{f}_{B_{1}^{rel}}$ and $\bm{f}_{p}-\bm{f}_{B_{2}^{rel}}$ capture the difference between the point and boxes. In addition, the box feature is obtained by averaging the point features within it. Note that before computing the box feature, we eliminate the background points in the box through Eq.~(\ref{eq:eq01}). To supervise the similarity learning, we use the cross-entropy loss to formulate pseudo label loss $L_{pl}$, which is defined as:
\begin{equation}
\setlength{\abovedisplayskip}{2pt}
\setlength{\belowdisplayskip}{2pt}
 L_{pl} = L_{CE}(\bm{A}, \bm{Y})
\label{eq:eq03}
\end{equation}
where $\bm{Y}$ is the label of the reliable point, which denotes the box it belongs to.

After that, we use the model trained on the reliable points (see $B_1^{rel}$ and $B_2^{rel}$ in Fig.~\ref{fig:figure02}) to predict the label of unreliable points (see $B_1^{un}$ and $B_2^{un}$ in Fig.~\ref{fig:figure02}). It is desired that the points located in the overlapping part of the box can be assigned to the corresponding instances, obtaining pseudo instance labels.

\subsection{Coarse-to-Fine Instance Segmentator}
\label{sec:method04}

The bounding box supervision provides coarse shape information for 3D instance segmentation. To obtain well-detailed instances, we present a coarse-to-fine instance segmentator that can recognize and segment the points in the point cloud into various instances. Specifically, we first employ the queries to learn from the whole point cloud to obtain coarse instances, and then these queries learn from the regions indicated by the coarse instances with Multi-level Attention Blocks to refine the instances.

\begin{figure}
  \centering
   \includegraphics[width=1.0\linewidth]{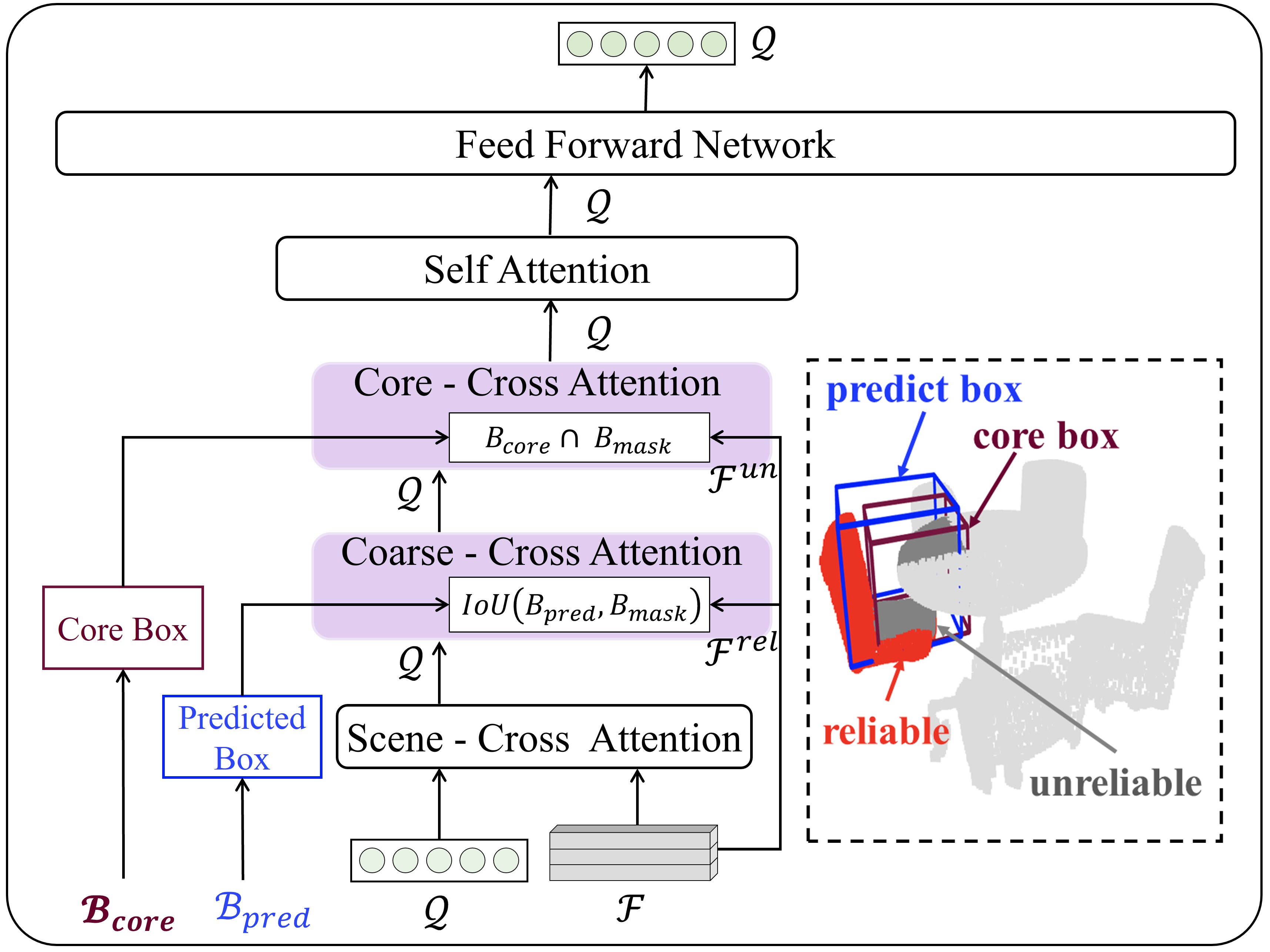}
   \vspace{-1.5em}
   \caption{\textbf{The details of Multi-level Attention Block.} The instance queries interact with the features of the whole scene, the coarse instance regions, and the instance core regions hierarchically. And the $B_{mask}$ are the boxes obtained from the predicted masks.}
   \label{fig:figure03}
   \vspace{-1.0em}
\end{figure}

\textbf{Coarse Instances Segmentation.}
Supposing one scene consists of $\mathit{q}$ objects, and each query represents one object and models the instance-related characteristics: the semantic category and the mask of each instance. In our method, these queries first interact with the backbone features of the whole point cloud to obtain coarse instances. Considering the point-wise masks of the coarse instances can not describe one object holistically and precisely, we predict another two boxes with coarse granularity and fine granularity, respectively. Then these two boxes combined with the scene-level features to refine the initial coarse instances with Multi-level Attention Blocks.

\textbf{Hierarchical Instances Refinement.}
To leverage the advantages of multi-granularity features from the point cloud and the segmented coarse instances, we introduce a novel decoder structure utilizing attentions to make the instance queries interact with granularity-varied features hierarchically. The decoder consists of a stack of six blocks. Each block includes a Multi-level Attention Block and an instance segmentation head illustrated in Fig.~\ref{fig:figure02}.



\textbf{Multi-level Attention Block.} The details of the Multi-level Attention Block are shown in Fig.~\ref{fig:figure03}. These instance queries first interact with point-wise backbone features globally, then interact with regions described by the coarse instance locally. 
The regions described by the coarse instance are divided into reliable and unreliable, which are depicted by the predicted box and the core box, respectively. To fully explore various granularity of shape information implied by the coarse instances and those two boxes, we first abstract the obtained coarse instance into a box, termed as $B_{mask}$. Then we explore the consistent information between $B_{mask}$ and the predicted box $B_{pred}$ to obtain features of reliable regions, which are described as follows:
\begin{equation}
\setlength{\abovedisplayskip}{2pt}
\setlength{\belowdisplayskip}{2pt}
 \bm{F}^{rel} = \sigma(\bm{F}, \bm{M} \odot e^{IoU(\bm{B}_{pred}, \bm{B}_{mask})})
  \label{eq:eq04}
\end{equation}
where $\sigma$ denotes the thresholding operation and $\odot$ denotes point-wise multiplication. In addition, $\bm{F}$ denotes the backbone features, $M$ denotes the predicted mask probabilities. $\bm{B}_{pred}$ and $\bm{B}_{mask}$ denote the predicted boxes and boxes decided by predicted masks, respectively.
So far, the instance queries have learned from the entire point cloud and the reliable regions, they can segment the majority of objects with easy contexts. To tackle the complex contexts, the instance queries learn from the unreliable regions described by the predicted core boxes. The predicted core boxes are scaled from the predicted boxes, aiming to represent the core regions of objects, and corresponding features are obtained as follows:
\begin{equation}
\setlength{\abovedisplayskip}{2pt}
\setlength{\belowdisplayskip}{2pt}
 \bm{F}^{un} = \sigma(\bm{F}, \bm{M} \odot (\bm{B}_{core} \cap \bm{B}_{mask}))
  \label{eq:eq05}
\end{equation}
where $\bm{B}_{core}$ denotes the predicted core boxes.
The instance queries have strong discriminative ability in such a global-to-local manner, then they are fed into a self-attention module and a feed-forward network to aggregate the learned information, which further boosts the performance.

\subsection{Bilateral Matching and Training Loss}
\label{sec:method05}

Our method makes bilateral matching between pseudo instances and predicted instances first and then utilizes the paired pseudo instances as ground truth $\mathbf{y}_{ins}$ to supervise the paired predicted instances $\textbf{x}_{ins}$.

\textbf{Bilateral matching.}
Following SPFormer \cite{sun2023superpoint} and ISBNet \cite{ngo2023isbnet}, our method utilizes the Hungarian method \cite{kuhn1955hungarian} to achieve the optimal assignment. We compute a pairwise matching cost $\mathbf{c}_{i,j}$ to evaluate the similarity of the $i^{th}$ predicted instance and the $j^{th}$ pseudo ground truth. And $\mathbf{c}^{i,j}$ is based on the classification probabilities and the binary masks, it is defined as:
\begin{equation}
\setlength{\abovedisplayskip}{2pt}
\setlength{\belowdisplayskip}{2pt}
    \begin{aligned}
     \mathbf{c}^{i,j} &=  \lambda_1 L_{CE}(\mathbf{x}^i_{cls}, \mathbf{y}^j_{cls}) \\
     & + \lambda_2 (L_{BCE}(\mathbf{x}^i_{mask}, \mathbf{y}^j_{mask})
     + L_{dice}(\mathbf{x}^i_{mask}, \mathbf{y}^j_{mask}))
      \label{eq:eq06}
    \end{aligned}
\end{equation}

\textbf{Training loss.}
Our training loss consists of two items, namely the loss of pseudo label generation $L_{pl}$ and the loss of instance segmentation $L_{seg}$, which is defined as:
\begin{equation}
\setlength{\abovedisplayskip}{2pt}
\setlength{\belowdisplayskip}{2pt}
 L = L_{pl} + L_{seg}
  \label{eq:eq07}
\end{equation}
As for the instance segmentation, except for supervising the predicted semantic classes and the binary masks, we also supervise the predicted bounding boxes and the core boxes utilized in the coarse-to-fine instance segmentator. The ground truth of the scale of the core box is based on the IoU score of the predicted masks and the pseudo-ground truth masks. The loss of instance segmentation is defined as:
\begin{equation}
\setlength{\abovedisplayskip}{2pt}
\setlength{\belowdisplayskip}{2pt}
    \begin{aligned}
     &L_{seg} = \lambda_1 L_{CE}(\mathbf{x}_{cls}, \mathbf{y}_{cls}) \\
     & + \lambda_2 (L_{BCE}(\mathbf{x}_{mask}, \mathbf{y}_{mask})
     + L_{dice}(\mathbf{x}_{mask}, \mathbf{y}_{mask})) \\
     & + \lambda_3 (L_{L_1}(\mathbf{x}_{box}, \mathbf{y}_{box})
     + L_{MSE}(\mathbf{x}_{core-box}, \mathbf{y}_{core-box}))
     \label{eq:eq08} 
    \end{aligned}
\end{equation}

\section{Experiments}
\label{sec:experiments}
\subsection{Experimental Setup}
\label{sec:experiments01}

\textbf{Datasets.}
We conduct experiments on two datasets, ScanNetV2 \cite{dai2017scannet} and S3DIS \cite{armeni20163d}. The ScanNetV2 provides splits containing 1201, 312, and 100 scenes for training, validation, and testing. Scenes are annotated with semantic labels covering 18 categories and instance labels. And the S3DIS dataset is another indoor dataset contains 272 scenes across six different areas, and it is annotated with 13 object categories. Following existing methods \cite{chibane2022box2mask, ngo2023gapro, lu2024bsnet}, the scenes of S3DIS Area 5 are used for validation while the others are used for training.

\textbf{Evaluation metrics.}
We use the mean average precision ($AP$) as the evaluation metric for both ScanNetV2 and S3DIS, which is utilized as a common evaluation metric in 3D instance segmentation. $AP$ is the averaged score with IoU thresholds from $95\%$ to $50\%$ with a step size of $5\%$. The mean average precision with IoU thresholds at 50\% and 25\% are used as evaluation metrics as well, which are denoted as $AP_{50}$ and $AP_{25}$, respectively.

\textbf{Implementation details.}
Our model is trained on a single RTX 3090 GPU, and we utilize SPFormer \cite{sun2023superpoint} and ISBNet \cite{ngo2023isbnet} as codebases for ScanNetV2 and S3DIS, respectively. 
We utilize the AdamW optimizer with a learning rate of $0.0002$ and weight decay at $0.05$ for the training of ScanNetV2. As for the weights of matching cost and losses, ($\lambda_1$, $\lambda_2$, $\lambda_3$) are set as ($0.5$, $1.0$, $0.5$). All the models are trained from scratch, and without specification, the hyper-parameters and the training details are kept the same as the \cite{sun2023superpoint} or \cite{ngo2023isbnet}. 
\begin{table}\footnotesize
   \caption{\textbf{Comparison on ScanNetV2.} ``Mask'' denotes using dense point-wise annotations, \textbf{$S_{0}$ and $S_{1}$ denote using accurate bounding box and sketchy bounding box annotations}, respectively. }
   \vspace{-1.0em}
  \label{tab:tab02}
  \centering
  \setlength{\tabcolsep}{0.5mm}
  \begin{tabular}{@{}lcccccc@{}@{}@{}@{}@{}@{}}
    \toprule
    \multirow{2}{*}{Method} & \multirow{2}{*}{Sup.} & \multirow{2}{*}{Venue} & \multicolumn{2}{c}{ScanNet Test} & \multicolumn{2}{c}{ScanNet Val} \\
    \cmidrule(r){4-5} \cmidrule(r){6-7}
     & & & $AP_{50}$ & $AP_{25}$ & $AP_{50}$ & $AP_{25}$ \\
    \midrule
    GraphCut \cite{hui2022learning} & \multirow{6}*{Mask} & NIPS22 & 73.2 & 83.2 & 69.1 & 79.3 \\
    ISBNet \cite{ngo2023isbnet} & ~ & CVPR23 & 76.3 & 84.5 & 73.1 & 82.5 \\
    SPFormer \cite{sun2023superpoint} & ~ & AAAI23 & 77.0 & 85.1 & 73.9 & 82.9 \\
    Mask3D \cite{schult2023mask3d} & ~ & ICRA23 & 78.0 & 87.0 & 73.7 & 83.5 \\
    MSTA3D \cite{tran2024msta3d} & ~ & ACM MM24 & 79.5 & 87.9 & 77.0 & 85.4 \\
    Spherical Mask \cite{shin2024spherical} & ~ & CVPR24 & 81.2 & - & 79.9 & 88.2 \\
    \midrule
    Box2Mask \cite{chibane2022box2mask} \ & \multirow{7}*{$S_{0}$} & ECCV22 & 67.7 & 80.3 & 59.7 & 71.8 \\
    WISGP+PointGroup \cite{du2023weakly} & ~ & WACV23 & - & - & 50.2 & 64.9 \\
    WISGP+SSTNet \cite{du2023weakly} & ~ & WACV23 &  - & - & 56.9 & 70.2 \\
    GaPro+SPFormer \cite{ngo2023gapro} & ~ & ICCV23 & 69.2 & 82.4 & 70.4 & 79.9 \\
    CIP-WPIS \cite{yu20243d} & ~ & WACV24 & - & - & 69.3 & 78.6 \\
    BSNet+SPFormer \cite{lu2024bsnet} & ~ & CVPR24 & - & - & \textbf{72.7} & 83.4 \\
    Sketchy-3DIS (ours) & ~ & - & \textbf{70.1} & \textbf{86.6} & 68.8 & \textbf{83.6} \\
    \midrule
    Box2Mask \cite{chibane2022box2mask} \ & \multirow{3}*{$S_{1}$} & ECCV22 & - & - & 52.4 & 67.5 \\
    GaPro+SPFormer \cite{ngo2023gapro} & ~ & ICCV23 & - & - & 53.5 & 72.2 \\
    Sketchy-3DIS (ours) & ~ & - & - & - & \textbf{65.8} & \textbf{83.1}\\
    \bottomrule
  \end{tabular}
  \vspace{-1.0em}
\end{table}

\subsection{Main Results}
\label{sec:experiments02}
\textbf{ScanNetV2.} In Table \ref{tab:tab02}, we compare our method and existing state-of-the-art methods on ScanNetV2 dataset. Attributed to the innovative design of Sketchy-3DIS, which achieves the pseudo labeling and instance segmenting simultaneously, our method achieves considerable performance on both the validation set and online test set of ScanNetV2. Notably, without multi-modal priors \cite{yu20243d} or synthesized scenes \cite{lu2024bsnet}, our method outperforms existing methods in terms of $AP_{25}$, which demonstrates the effectiveness of our method. Additionally, the comparable performances to fully-supervised methods demonstrate that our method can play a significant role in autonomous driving with high efficacy.

\begin{table}\small
   \caption{\textbf{Comparison on S3DIS Area 5}. ``Mask'' denotes using dense point-wise annotations, \textbf{$S_{0}$ and $S_{1}$ denote using accurate bounding box and sketchy bounding box annotations}, respectively. ``$^{*}$'' denotes the results are reproduced by GaPro \cite{ngo2023gapro}. 
   }
   \vspace{-1.0em}
  \label{tab:tab03}
  \centering
  \setlength{\tabcolsep}{2mm}
  \begin{tabular}{@{}lcccc@{}@{}@{}@{}@{}}
    \toprule
    Method & Sup. & Venue & $AP$ & $AP_{50}$ \\
    \midrule
    SoftGroup \cite{vu2022softgroup} & \multirow{6}*{Mask} & CVPR22 & 51.6 & 66.1 \\
    GraphCut \cite{hui2022learning} & ~ & NIPS22 & 54.1 & 66.4 \\
    ISBNet \cite{ngo2023isbnet} & ~ & CVPR23 & 54.0 & 65.8 \\
    Mask3D \cite{schult2023mask3d} & ~ & ICRA23 & 56.6 & 68.4 \\
    MSTA3D \cite{tran2024msta3d} & ~ & ACM MM24 & - & 70.0  \\
    Spherical Mask \cite{shin2024spherical} & ~ & CVPR24 & 60.5 & 72.3 \\
    \midrule
    Box2Mask$^{*}$ \cite{chibane2022box2mask} & \multirow{6}*{$S_{0}$} & ECCV22 & 43.6 & 54.6 \\
    WISGP+PointGroup \cite{du2023weakly} & ~ & WACV23 & 33.5 & 48.6 \\
    WISGP+SSTNet \cite{du2023weakly} & ~ & WACV23 & 37.2 & 51.0 \\
    GaPro+ISBNet \cite{ngo2023gapro} & ~ & ICCV23 & 50.5 & 61.2 \\
    BSNet+ISBNet \cite{lu2024bsnet} & ~ & CVPR24 & 53.0 & 64.3 \\
    Sketchy-3DIS (ours) & ~ & - &  \textbf{53.4} & \textbf{69.1} \\
    \midrule
    Sketchy-3DIS (ours) & $S_{1}$ & - &  \textbf{50.7} & \textbf{64.6} \\
    \bottomrule
  \end{tabular}
  \vspace{-1.0em}
\end{table}

\begin{figure*}[t]
  \centering
   \includegraphics[width=0.88\linewidth]{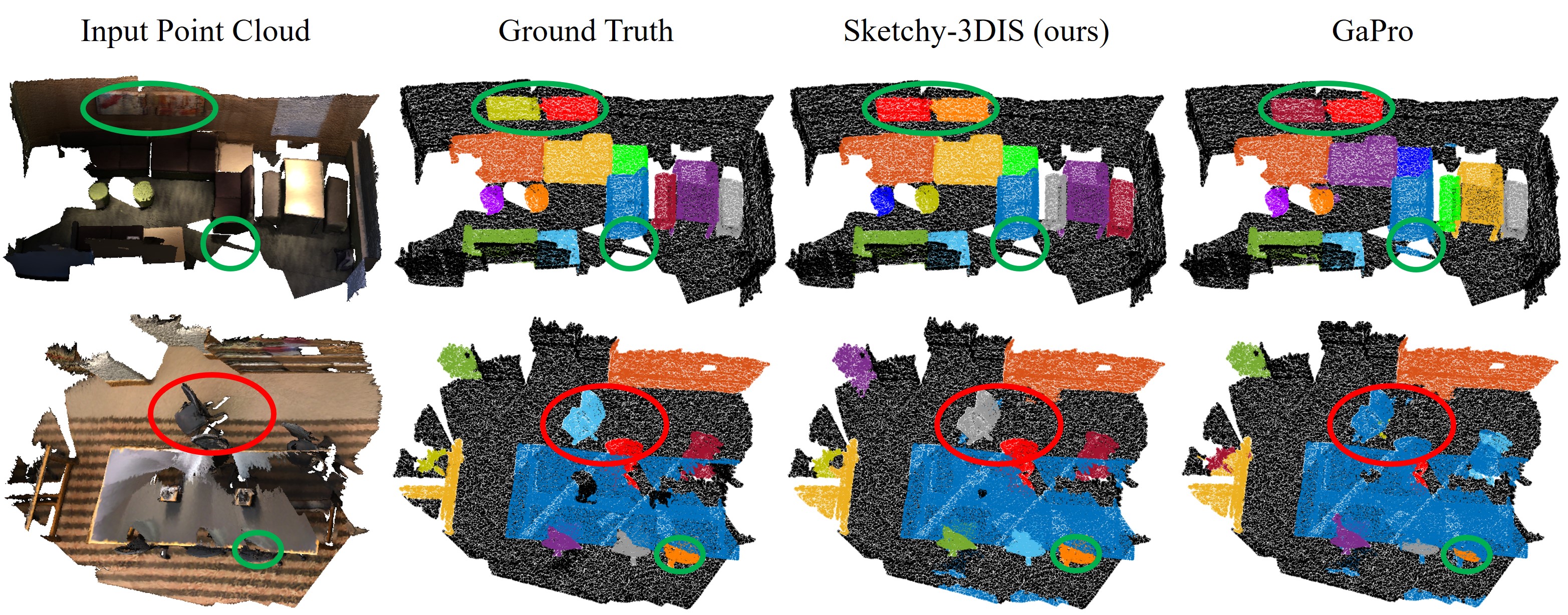}
   \vspace{-0.8em}
   \caption{\textbf{Visualization comparison of pseudo labels on the ScanNetV2 training set.} The black denotes the background points and other colors denote different objects. The green and red cycles highlight the key regions.}
   \vspace{-1.0em}
   \label{fig:figure04}
\end{figure*}

\textbf{S3DIS.} In Table \ref{tab:tab03}, we make comparisons between existing methods and our proposed Sketchy-3DIS on S3DIS Area 5. The consistent advantages demonstrate the strong generalization ability of our proposed SKeychy-3DIS. Notably, our method exceeds the fully-supervised baseline~\cite{ngo2023isbnet} with $\mathbf{+3.3}$ in $AP_{50}$, which verifies that our proposed Sketchy-3DIS can make full use of instance indications provided by the sketchy bounding boxes.


\textbf{Quality comparisons of pseudo labels.} We compare the qualities of the generated pseudo labels of GaPro \cite{ngo2023gapro} and ours under $S_{1}$ sketchy bounding boxes. The qualitative comparisons are present in Fig.\ref{fig:figure04}, which illustrates that our method generates more accurate pseudo labels than GaPro not only in details (highlighted by the green cycles) but also in distinguishing the objects belonging to the same categories (highlighted by the red cycles). Additionally, compared with the ground truth, our method generates comparable pseudo labels as well.

\subsection{Ablation Studies and Analysis}
\label{sec:experiments04}
We conduct comprehensive ablation studies to justify the effective design of our method. All these ablation experiments are conducted on the ScanNetV2 validation set unless otherwise stated.

\begin{table}\footnotesize
   \caption{\textbf{Robustness under various sketchy bounding boxes.} $S_{0}$ to $S_{4}$ denote the sketchy degree of the annotated bounding boxes.}
   \vspace{-1.0em}
  \label{tab:tab04}
  \centering
  \setlength{\tabcolsep}{4.5mm}
  \begin{tabular}{@{}lcccc@{}@{}@{}@{}}
    \toprule
    \multirow{2}{*}{Sketchy Degree} & \multicolumn{2}{c}{ScanNetV2 Val} & \multicolumn{2}{c}{S3DIS Area 5} \\
    \cmidrule(r){2-3} \cmidrule(r){4-5}
     & $AP_{50}$ & $AP_{25}$ & $AP_{50}$ & $AP_{25}$ \\
    \midrule
    $S_{4}$ & 62.5 & 80.1 & 50.9 & 63.6 \\
    $S_{3}$ & 62.9 & 81.4 & 52.3 & 66.6 \\
    $S_{2}$ & 63.7 & 82.1 & 61.2 & 72.7 \\
    $S_{1}$ & 65.8 & 83.1 & 64.6 & 73.9 \\
    $S_{0}$ & \textbf{68.8} & \textbf{83.6} & \textbf{69.1} & \textbf{77.5} \\
    \bottomrule
  \end{tabular}
\end{table}

\begin{table}
  \caption{\textbf{Effectiveness of pseudo labeler and instance segmentator.} UnScene3D \cite{rozenberszki2024unscene3d} is an sota unsupervised method. ``disjoint'' denotes training the 3D UNet and pseudo labeler first and then freeze them to train instance segmentator while ``joint'' denotes training the pseudo labeler and the instance segmentator jointly.}
  \vspace{-1.0em}
  \label{tab:tab05}
  \centering
  \setlength{\tabcolsep}{3.8mm}
  \begin{tabular}{@{}lcccc@{}@{}@{}@{}}
    \toprule
    Method & $AP$ & $AP_{50}$ & $AP_{25}$ \\
    \midrule
    UnScene3D \cite{rozenberszki2024unscene3d} & 21.4 & 40.3 & 52.6 \\
    SKetchy-3DIS (disjoint) & 45.3 & 60.4 & 70.0 \\
    SKetchy-3DIS (joint) & \textbf{53.4} & \textbf{69.1} & \textbf{77.5} \\
    \bottomrule
  \end{tabular}
  \vspace{-1.0em}
\end{table}

\textbf{Robustness under various sketchy bounding boxes.} In Table \ref{tab:tab04}, we evaluate our method under four kinds of various sketchy bounding boxes illustrated in Fig.\ref{fig:figure10} on both ScanNetV2 validation set and S3DIS Area 5. It can be observed that the performance of our method deteriorates slightly as the increasing of the boxes' sketchy degree, which verifies the robustness of our proposed Sketchy-3DIS.

\textbf{Effectiveness of labeler and segmentator.} In Table \ref{tab:tab05}, experiments are conducted on the S3DIS Area 5 to verify the effectiveness of our designed adaptive box-to-point pseudo labeler and coarse-to-fine instance segmentator. We choose the state-of-the-art unsupervised method UnScene3D \cite{rozenberszki2024unscene3d} for 3D instance segmentation as the baseline. The experimental results verify that our designed pseudo labeler can generate high-quality pseudo labels, and combined with our designed instance segmentator, our method achieves significant performance.

\begin{table}
  \caption{\textbf{Effectiveness of components in pseudo labeler.} ``Partition'' denotes obtaining point-wise labels by box-to-point spatial correlations, ``Assign'' denotes assigning the points within over-lapped boxes, and ``Similarity'' denotes filtering the background points within boxes according to the point-to-box similarities.}
  \vspace{-1.0em}
  \label{tab:tab06}
  \centering
  \setlength{\tabcolsep}{2.5mm}
  \begin{tabular}{@{}lccccc@{}@{}@{}@{}@{}}
    \toprule
    Partition & Assign & Similarity & $AP$ & $AP_{50}$ & $AP_{25}$\\
    \midrule
     & &  & 15.9 & 32.2 & 58.5 \\
    \checkmark & & & 41.8 & 64.8 & 72.3 \\
    \checkmark & \checkmark & & 45.2 & 67.3 & 83.4 \\
    \checkmark & \checkmark & \checkmark & \textbf{46.0} & \textbf{68.8} & \textbf{83.6} \\
    \bottomrule
  \end{tabular}
\end{table}

\begin{table}[h]
\caption{\textbf{Attentions in Multi-level Attention Block.} ``Scene'', ``Coarse'', and ``Core'' denote interacting with the features of the whole scene, the coarse instance regions, and the core regions of the instance, respectively. And ``Self'' denotes augmenting by the learned instance query itself.}
\vspace{-1.0em}
\label{tab:tab07}
\centering
\setlength{\tabcolsep}{2.5mm}
    \begin{tabular}{@{}ccccccc@{}@{}@{}@{}@{}@{}}
        \toprule
         Scene & Coarse & Core & Self & $AP$ & $AP_{50}$ & $AP_{25}$ \\
        \midrule
         & & & & 41.3 & 64.1 & 79.8\\
        \checkmark & & & \checkmark & 44.6 & 67.0 & 82.1 \\
        \checkmark & \checkmark & & \checkmark & 44.7 & 67.1 & 82.9 \\
        \checkmark & \checkmark & \checkmark & \checkmark & \textbf{46.0} & \textbf{68.8} & \textbf{83.6} \\
        \bottomrule
    \end{tabular}
\end{table}

\begin{table}[h]
\caption{\textbf{Effectiveness of instance segmentation losses.} $L_{box}$, $L_{core-box}$, and $L_{mask}$ denote the supervisions of the predicted instance box, instance core box, and instance mask, respectively. }
\vspace{-1.0em}
\label{tab:tab08}
\centering
\setlength{\tabcolsep}{3.0mm}
    \begin{tabular}{@{}cccccc@{}@{}@{}@{}@{}}
        \toprule
        \rule{0pt}{10.5pt} 
       $L_{box}$ & $L_{core-box}$ & $L_{mask}$ & $AP$ & $AP_{50}$ & $AP_{25}$ \\
        \midrule
        \rule{0pt}{9pt} 
        \checkmark & & & 45.2 & 66.8 & 82.3 \\
        \rule{0pt}{9pt} 
        \checkmark & \checkmark & & 45.4 & 67.7 & 83.0 \\
        \rule{0pt}{9pt} 
        \checkmark &  & \checkmark & 45.0 & 67.8 & 83.0 \\
        \rule{0pt}{8.8pt} 
        \checkmark & \checkmark & \checkmark & \textbf{46.0} & \textbf{68.8} & \textbf{83.6} \\
        \bottomrule
    \end{tabular}
    \vspace{-1.0em}
\end{table}


\textbf{Effectiveness of components in pseudo labeler.} In Table \ref{tab:tab06}, we evaluate the effectiveness of various components in our designed pseudo labeler. The first row is the results of the state-of-the-art unsupervised method \cite{rozenberszki2024unscene3d}, and the comparison between the last two rows verifies that the learned point-to-box similarity and point-to-instance assignments can improve the performance by filtering the background points within boxes and assigning the points within overlapped boxes.

\textbf{Effectiveness of various attentions in Multi-level Attention Block.} In Table \ref{tab:tab07}, we conduct ablation studies on various combinations of attention layers utilized in Multi-level Attention Block. The best performance is obtained with the combination of the four attention modules. These experimental results verify that our proposed Multi-level Attention Block can explore the local contexts adaptively and aggregates them, which boosts the segmentation performance progressively.

\textbf{Effectiveness of instance segmentation losses.} In Table \ref{tab:tab08}, we compare the performances of the losses of the predicted boxes, core boxes, and masks. The results show that the best performance is obtained with the combination of the above three kinds of losses, combined with our coarse-to-fine instance segmentator, they achieve inspiring performance.

\begin{figure}[t]
  \centering
   \includegraphics[width=1.0\linewidth]{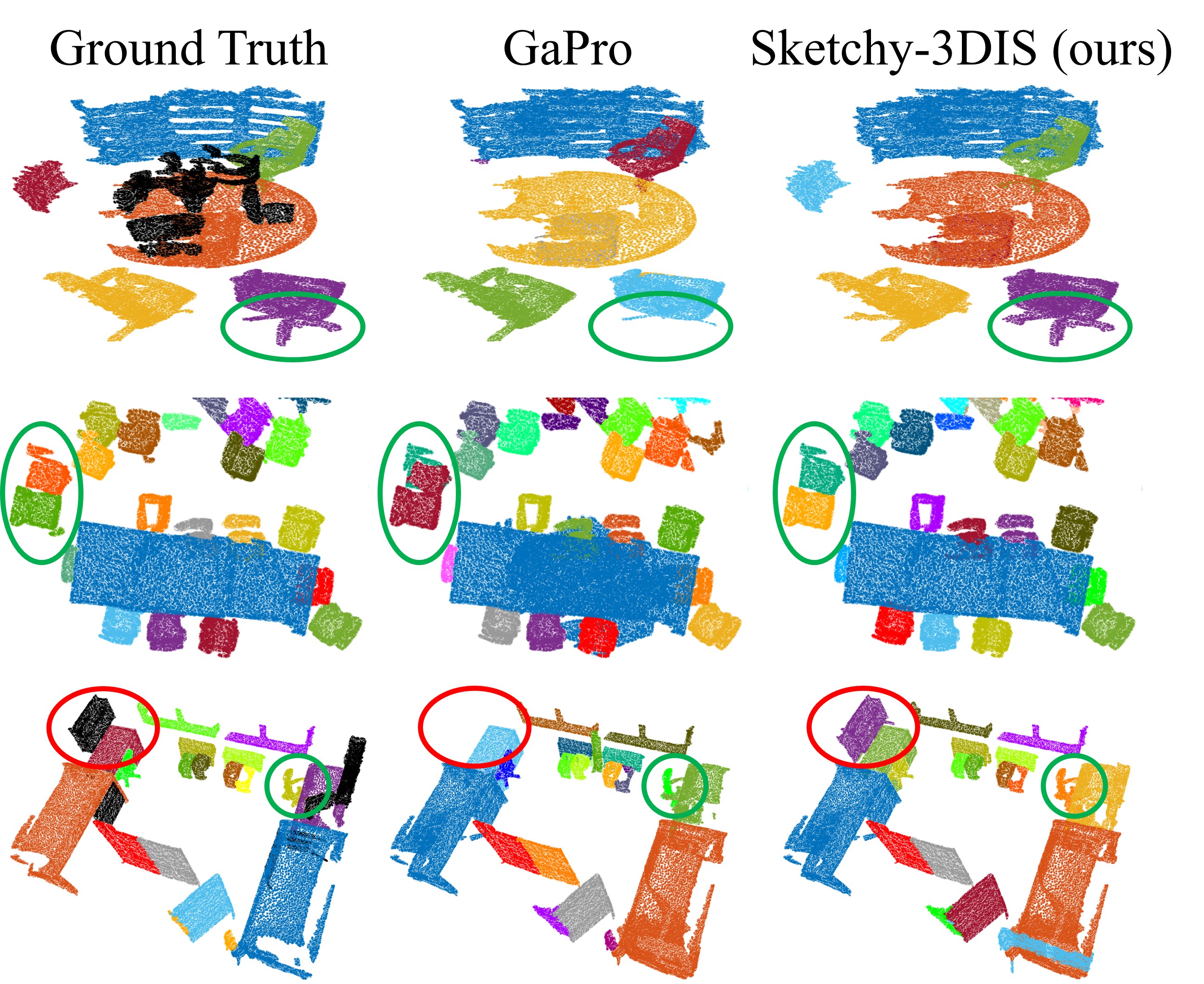}
   \caption{\textbf{Qualitative results on ScanNetV2 validation set.} The black denotes the background points and other colors denote different objects. The cycles highlight the key regions.}
   \label{fig:figure06}
   \vspace{-1.0em}
\end{figure}

\textbf{Qualitative results of 3D instance segmentation.}
Some visualized samples of predicted instances for the ScanNetV2 validation set are shown in Fig.\ref{fig:figure06}. It can be observed that our method shows consistent effectiveness in segmenting objects for easy and complex scenes. Even if there are many objects belonging to the same category and they are close to each other, our method demonstrates considerable performance with the sketchy bounding box annotation. Notably, our method can adaptively segment the objects that are ignored in the ground truth such as the cabinets highlighted in red cycles, which demonstrates the strong recognition capability of our method. 

\section{Conclusion}
\label{sec:conclusion}
In this work, we propose a query-based framework for weakly supervised 3D instance segmentation, Sketchy-3DIS, which is more robust and effective than existing bounding-box supervised methods. Sketchy-3DIS adaptively generates pseudo per-point labels from sketchy boxes and predicts instances in a coarse-to-fine manner. Extensive experiments on ScanNetV2 and S3DIS benchmarks show that our method achieves leading performance, and even outperforms some fully-supervised methods. However, our explored experiments demonstrate that the performance of our method degraded severely when the annotated sketchy bounding boxes are immensely inaccurate, exploration for this issue may be the future study orientation.



{
    \small
    \bibliographystyle{ieeenat_fullname}
    \bibliography{main}
}

\end{document}